# SUPERVISED CONTRASTIVE LEARNING AS MULTI-OBJECTIVE OPTIMIZATION FOR FINE-TUNING LARGE PRE-TRAINED LANGUAGE MODELS

**Youness Moukafih**[1,2], **Mounir Ghogho**[1], **Kamel Smaili**[2]

1:TICLab, College of Engineering and Architecture, Université Internationale de Rabat, Morocco

2:LORIA/INRIA-Lorraine 615 rue du Jardin Botanique, BP 101, F-54600 Villers-16s-Nancy, France

`youness.moukafih@uir.ac.ma`

September 28, 2022

## ABSTRACT

Recently, Supervised Contrastive Learning (SCL) has been shown to achieve excellent performance in most classification tasks. In SCL, a neural network is trained to optimize two objectives: pull an anchor and positive samples together in the embedding space, and push the anchor apart from the negatives. However, these two different objectives may conflict, requiring trade-offs between them during optimization. In this work, we formulate the SCL problem as a Multi-Objective Optimization problem for the fine-tuning phase of RoBERTa language model. Two methods are utilized to solve the optimization problem: (i) the linear scalarization (LS) method, which minimizes a weighted linear combination of pertask losses; and (ii) the Exact Pareto Optimal (EPO) method which finds the intersection of the Pareto front with a given preference vector. We evaluate our approach on several GLUE benchmark tasks, without using data augmentations, memory banks, or generating adversarial examples. The empirical results show that the proposed learning strategy significantly outperforms a strong competitive contrastive learning baseline.

## 1 Introduction

Recently, contrastive learning has achieved state-of-the-art performance in various artificial intelligent applications, including Natural Language Processing (NLP) Gunel et al. [2020], Moukafih et al. [2022], Giorgi et al. [2020], Computer Vision (CV) Chen et al. [2020] and graph representation learning Hafidi et al. [2020, 2022]. Many approaches have been proposed to learn high-quality representations by minimizing a contrastive loss. The main common idea behind these approaches is as follows: train an encoder, a neural network, to increase both intra-class compactness and inter-class separability in the embedding space. In other words, the goal is to train a model to optimize two objectives: embed examples belonging to the same class close to each other, and embed examples from different classes further apart. Contrastive learning has been used in both self-supervised and supervised Learning settings. In the former, positive pairs are created by performing data augmentation methods, while the negatives are formed by the anchor and randomly chosen examples from the same mini-batch. In the latter, label information is leveraged by considering samples belonging to the same class as positive examples for each other, while negatives are samples from the remaining classes. The training in both self-supervised/Supervised Contrastive Learning is usually done in two steps: a pretraining step where a contrastive loss is minimized using the encoder's normalized representations, followed by a conventional training step where a simple linear model having as input the learnt representation is trained using the cross-entropy loss. In Gunel et al. [2020], the authors proposed to combine the cross-entropy loss and a SCL Khosla et al. [2020] for fine-tuning a large language model. Their method improves the performance on several NLP classification tasks from the GLUE benchmark over fine-tuning RoBERTa-large model using cross-entropy loss especially in the few-shot learning setting.

A variety of contrastive losses have been proposed for optimizing the two objectives mentioned above Gao et al. [2021], Khosla et al. [2020], Moukafih et al. [2022]. However, these objectives may conflict with each other, thus requiring trade-offs between them during the optimization process. The main objective of this work is to investigate this issue and devise new solutions using the Multi-Objective Optimization (MMO) framework. In MOO, the aim is to find a set of feasible solutions that are not-dominated by other feasible solutions, i.e., no objective value can be improved further without degrading some other objectives. This is referred to as the Pareto front. MOO has a wide variety of applications in machine learning Désidéri [2012], Sener and Koltun [2018], Navon et al. [2020].

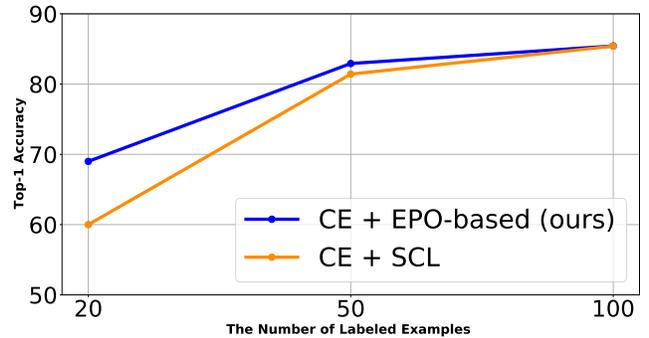

Figure 1: Our method (CE + EPO-based) consistently outperforms the method (CE + SCL) proposed in Gunel et al. [2020] on SST-2 dataset. We show top-1 accuracy on test set with 20, 50, 100 labeled training examples.

Several algorithms exist for MOO. The most straight-forward approach to MOO is the linear scalarization (LS) which minimizes the weighted sum of the objective functions given a preference vector $r$. The preference vector, consisting of the weighting parameters, represents a single trade-off between objectives. LS has a major limitation: the Pareto optimal solution cannot be obtained for all preference vectors when the objectives are non-convex (Figure 2a). However, LS tends to work well in practice. In Désidéri [2012], the authors proposed the multiple-gradient descent algorithm (MGDA), which uses gradient-based optimization to find one solution on the Pareto front. However, different solutions are found for different initializations and the preference vector is not taken into account. In Sener and Koltun [2018], the authors proposed a multi-task learning (MTL) algorithm from the MOO perspective that scales to high-dimensional problems. Instead of the uniform weight strategy, they used the MGDA algorithm to determine the optimal weights to obtain a solution on the Pareto front. Their method, however, finds only one single arbitrary Pareto-optimal solution. In , the authors proposed Pareto multi-task learning (Pareto MTL), a method that splits the objective space into separate cones given a set of preference rays, and returns a solution per cone. Their approach is capable of finding several points on the Pareto front; However, it scales poorly with the number of cones and does not converge to the exact desired ray on the Pareto front (see Figure2b. Recently, Mahapatra and Rajan [2021] proposed a new approach to find the Exact Pareto Optimal (EPO) solution in the objective space (the intersection of the Pareto front with a given preference ray) as illustrated in Figure 2c.

The technical novelty of our work resides in the formulation of the SCL problem as a MOO problem. To the best of our knowledge, this is the first work on fine-tuning a large language model by minimizing two objective functions using MOO. The first objective function encourages the encoder to maximize the agreement between a cluster of points belonging to the same class in the latent space, while the second guides the encoder to represent sentences form different classes far away from each other in the latent space. We address the MOO problem using both LS and EPO methods.

We show the merit of our approach on multiple datasets form the GLUE natural language understanding benchmark. We evaluate the method on both few-shot (20, 50, 100 labeled examples) learning as well as the full dataset training settings. The experiments show the superiority of our approach over two very strong baselines that fine-tune RoBERTa-base language model using CE only and CE + SCL objective respectively (see figure 1). For instance, on the SST-2 dataset, when the number of annotated examples is $N = 20$, our EPO-based method outperforms CE and CE + SCL-based methods by 9.24 and 8.62 percentage points respectively. Our approach outperforms the two baselines on sentence-pair classification tasks; over 2 percentage points improvement on the MNLI dataset when the number of labeled examples is 100.

Our contributions can be summarized as follows:

- we formulate the SCL problem as a MMO problem;

- we propose a method to fine-tune pre-trained language models by using SCL and MOO (here we focus on RoBERTa-base language models);

- we adapt two well-known MOO approaches to solve several downstream tasks from the GLUE benchmark.

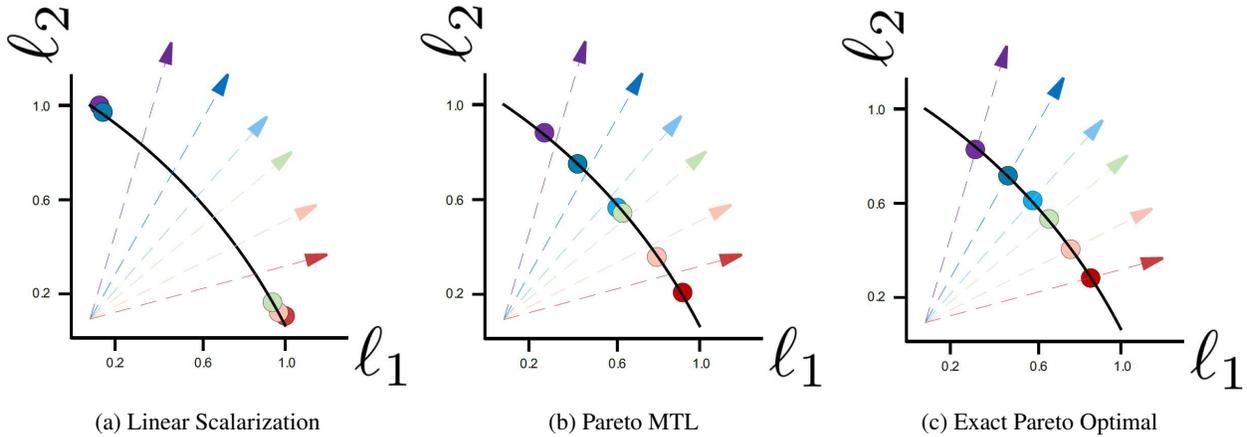

(a) Linear Scalarization      (b) Pareto MTL      (c) Exact Pareto Optimal

Figure 2: Pareto front (solid curve) for two non-convex loss functions $\ell_1, \ell_2$, and solutions on the Pareto front for different preferences (dashed rays) obtained by (a) Linear Scalarization, (b) Pareto MTL-based method, and (c) Exact Pareto Optimal search.

## 2 Background and Literature Review

### 2.1 Multi-Objective Optimization

MOO aims to find a set of non-dominated solutions in the Pareto front for multiple and possibly contrasting objectives. Conventional approaches follow genetic algorithms that search for populations of Pareto optimal solutions Deb [2001], Coello [2006]. However, the main disadvantage of these algorithms is that they do not scale to large neural network architectures. To address this issue, Désidéri [2012] proposed MGDA to find an optimal solution by finding a descent direction that decreases all objectives using multi-objective Karush-Kuhn-Tucker (KKT) conditions. The authors of Sener and Koltun [2018] succeeded to treat MTL as a MOO problem by using the MGDA with the Franke-Wolfe optimizer to find one non-dominated solution in the objective space. However, these approaches do not take into consideration the preferences and find only a single arbitrary Pareto optimal solution. Follow-up methods extended the idea of finding Pareto-optimal solutions by aligning solutions according to preference vectors Lin et al. [2019], Mahapatra and Rajan [2021]. However, this approach has the following limitation: a separate model has to be trained for each point on the front. To address this problem, the authors of Navon et al. [2020] proposed to learn the entire Pareto front using the hyper-network which outputs a preference-dictated prediction model.

### 2.2 Self-supervised contrastive learning

Self-Supervised Learning (SSL) is a prominent area of research that ventured out of unsupervised representation learning. In this approach, pretext tasks are used to evaluate representations on the basis of intrinsic properties of the data. Self-supervised contrastive learning (SSCL) is one of the approaches to learn representation using unlabeled data. This tries to discriminate between two types of examples, a) positive examples and b) negative examples given an anchor. Since in SSL the labels are not available, the data augmentation techniques are performed to create positive pairs (e.g. Rotation and Cropping in the case of images), while the negatives are taken from the same mini-batch as the anchor Chen et al. [2020], He et al. [2020], Tian et al. [2020], Grill et al. [2020]. There have also been some recent works that explored methods and techniques for hard negative mining for contrastive learning Hafidi et al. [2022], Kalantidis et al. [2020]. Recognizing the importance of using a large number of negatives for contrastive loss-based representation learning, various recent approaches use memory banks to store the learnt representations He et al. [2020].

Inspired by the effectiveness of self-supervised contrastive learning in computer vision domain, various approaches have extended this method to learn representations of graph structured data You et al. [2020], Hafidi et al. [2022] as well as textual data representation Wu et al. [2020]. In Yan et al. [2021], the authors propose to enhance the native derived sentence representations by maximizing the agreement between a given anchor and the augmented positive pair by applying different data augmentation methods such as word shuffling and cropping. Their method improved the vanilla BERT on the semantic textual similarity (STS) tasks. More recently, Kim et al. [2021] advanced the state-of-the-art of BERT's representation even further without using any data augmentation method. The core idea behind their work is to recycle intermediate BERT's hidden representations as positive samples to which the final sentence embedding should be close. In Gao et al. [2021], the authors present SimCSE framework that achieves state-of-the-art results on the STS task; the dropout function is considered as a minimal data augmentation method. More formally, by passing the same sentence to the pre-trained encoder twice, two different embeddings are obtained because of the random dropout; these two embeddings are considered positive pairs, while the negatives are formed by the remaining sentences from the same mini-batch.

## 2.3 Supervised contrastive learning

Recent studies have extended SSL learning to the fully supervised framework by using tag information to learn representations. In fully-supervised contrastive learninig, positive pairs are elements from the same class as the anchor and negatives are examples from the remaining classes. In the computer vision domain, state-of-the-art results were obtained by a method based on this approach, called SupCon Khosla et al. [2020]). In that paper, the authors propose to maximize the average similarity between a cluster of points with the same class label, while maximizing the distance between examples from the other classes within the same mini-batch. More recently, another work in Supervised contrastive learning has been proposed for text representations Moukafih et al. [2022]. The method is similar to SupCon, except that it maximizes the average similarity between the anchor and negative examples. This was shown to achieve better results on text classification tasks than SupCon and the commonly used cross-entropy loss-based methods. Recently, in Gunel et al. [2020] the authors fine-tuned RoBERTa-large by minimizing a novel objective that includes a supervised contrastive learning term. Their proposed method outperforms significantly the vanilla RoBERTa-large model in the few-shot learning settings. However, in the full dataset training setting, the difference in performance is insignificant. In Gao et al. [2021], a supervised contrastive learning setting is also developed by pre-training the representation of BERT using natural language inference dataset (3 classes are involved, namely entailment, neutral, and contradiction). The entailment pairs are considered as positive instances, while the contradiction pairs as hard negatives.

## 3 The proposed approach

We propose to fine-tune the pre-traind RoBERTa language model by minimizing an objective function that combines the well-known cross-entropy and a supervised contrastive loss. The technical novelty of this work is to formulate the contrastive loss optimization as a MOO problem. We utilize two well known approaches to solve the MOO problem namely, the linear scalarization and the exact pareto optimal method Mahapatra and Rajan [2021].

Similar to Gunel et al. [2020], we fine-tune RoBERTa on single sentence and sentence-pair classification tasks. For single sentence classification tasks, given an input sentence $S_i$, RoBERTa language model generates a sequence of token embeddings prepended with the special $[CLS]$ token, $\bar{S}_i = [[CLS], r_1, r_2, \ldots r_L, [EOS]]$, where $[EOS]$ denotes end of sentence token and $L$ denotes the max-length. For sentence-pair classification on the other hand, each instance is a concatenation of two sequences of tokens $[[\bar{CLS}], r_1, r_2, \ldots r_L, [SEP], z_1, z_2, \ldots z_K, [EOS]]$, where $[SEP]$ denotes the separator token and $L + K$ is the max-length. Following common practice for fine-tuning pre-trained language models for classification Devlin et al. [2018], Gunel et al. [2020], we consider the $[CLS]$ token to be the final representation of the input data.

### 3.1 Preliminaries

Let us denote a labeled dataset as $\mathcal{D} = \{(\boldsymbol{x}_i, y_i)\}_i$, where $\boldsymbol{x}_i \in \mathcal{X}$ represents the $i^{th}$ instance of the dataset and $y_i \in \mathcal{Y} = \{1, ..., C\}$ is its label. We train an encoder function (a neural network) $f_{\boldsymbol{\theta}} : \mathcal{X} \longrightarrow \mathcal{Y}$ parameterized by $\boldsymbol{\theta}$.

The goal of MOO is to jointly minimize $m$ non-negative objective functions. In our setting, each objective function is the expectation of the individual (or mini-batch) loss, $\bar{\ell}_j$, over labeled instances of $\boldsymbol{x}$ and $y$, randomly sampled from the data distribution $\mathcal{P}_D$:

$$\ell_j(\boldsymbol{\theta}) = \mathbb{E}_{(\boldsymbol{x}, y) \sim \mathcal{P}_D} [\bar{\ell}_j (y, f_{\boldsymbol{\theta}}(\boldsymbol{x}))] \tag{1}$$

where $j \in \{1, \ldots, m\}$. The goal of MOO is to find Pareto optimal solutions.

We define a partial ordering on the objective space by $\boldsymbol{\ell}(\boldsymbol{\theta}) \preceq \boldsymbol{\ell}(\boldsymbol{\theta}')$, where $\boldsymbol{\ell}(\boldsymbol{\theta}) = [\ell_1(\boldsymbol{\theta}), \ldots, \ell_m(\boldsymbol{\theta})]^T \in \mathbb{R}_+^m$, if for all $j \in \{1, \ldots, m\}, \ell_j(\boldsymbol{\theta}) \leq \ell_j(\boldsymbol{\theta}')$; for strict inequality, i.e. $\boldsymbol{\ell}(\boldsymbol{\theta}) \prec \boldsymbol{\ell}(\boldsymbol{\theta}')$, we have $\ell_j(\boldsymbol{\theta}) < \ell_j(\boldsymbol{\theta}')$ for some of the values of $j$.

**Definition** (Pareto dominance). A solution $\boldsymbol{\theta}_1$ dominates a solution $\boldsymbol{\theta}_2$ if $\boldsymbol{\ell}(\boldsymbol{\theta}_1) \prec \boldsymbol{\ell}(\boldsymbol{\theta}_2)$. In other words, $\boldsymbol{\theta}_1$ is not worse than $\boldsymbol{\theta}_2$ on any objective, and $\boldsymbol{\theta}_1$ is better than $\boldsymbol{\theta}_2$ on at least one objective, i.e.: $\exists q \in \{1, \ldots, m\}$ s.t $\ell_q(\boldsymbol{\theta}_1) < \ell_q(\boldsymbol{\theta}_2)$. A point that is not dominated by any other point is called Pareto optimal solution. The set of all Pareto optimal solutions is called Pareto set, denoted by, $\mathcal{P}_{\boldsymbol{\theta}}$, and its image is called the Pareto front $\mathcal{P}_{\boldsymbol{\ell}} = \{\boldsymbol{\ell}(\boldsymbol{\theta})\}_{\boldsymbol{\theta} \in \mathcal{P}_{\boldsymbol{\theta}}}$.

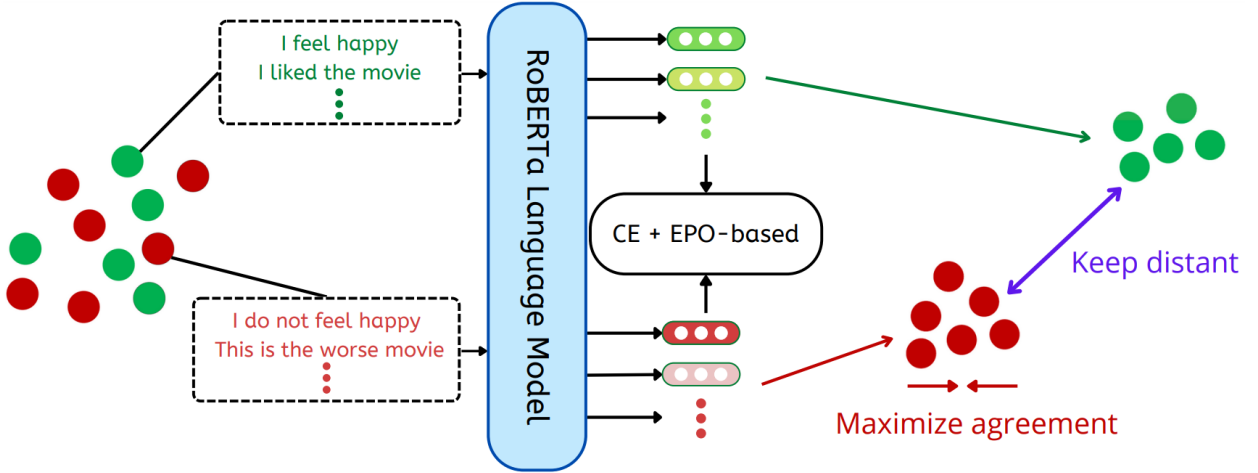

Figure 3: The general framework of our proposed approach

## 3.2 The proposed training strategy

We first define some needed notations, formulate the learning problem of interest, and then provide details of the proposed solution. Let $\mathcal{S}^k = \{(\boldsymbol{x}_i, y_i) | y_i = k\}$ denotes the set of all samples belonging to class $k$ within the corpus $\mathcal{D}$. Let $\mathcal{B}_k$ be a mini-batch of $N_k$ examples randomly sampled from $\mathcal{S}^k$. Let $\mathcal{H}_k = f_{\boldsymbol{\theta}}(\mathcal{B}_k) \in \mathbb{R}^{N_k \times d}$ be the $\ell_2$ normalization of the highest level representation of the neural network:

$$\mathcal{H}_k = \begin{bmatrix} \boldsymbol{h}_1^\top \\ \boldsymbol{h}_2^\top \\ \vdots \\ \boldsymbol{h}_{|\mathcal{B}_k|}^\top \end{bmatrix} \in \mathbb{R}^{N_k \times d}$$

where $\boldsymbol{h}_j$ is the embedding vector corresponding to instance $j$, whose dimension is denoted by $d$, and $|\cdot|$ denotes the cardinality operator.

Similar to Gunel et al. [2020], Khosla et al. [2020], our main goal is to maximize the similarity between points that belong to the same class and minimizing the similarity between elements of different classes. We now proceed with the formulation of the two objective functions. First, we define the following matrices:

$$\mathcal{M}^{(k)} = \mathcal{H}_k \mathcal{H}_k^T \in \mathbb{R}^{N_k \times N_k},$$

$$\mathcal{N}^{(k)} = [\mathcal{H}_k \mathcal{H}_{k'}^T]_{k' \in \mathcal{Y}, k' \neq k} \in \mathbb{R}^{N_k \times \overline{N_k}}$$

where $\overline{N_k} = \sum_{k' \neq k} N_k$ and $[.]$ denotes the horizontal concatenation operator. Matrices $\mathcal{M}^{(k)}$, with $k = 1, \cdots, C$, contain the intra-class similarities, whereas matrices $\mathcal{N}^{(k)}$ contain inter-class similarities.

In this work, we aim to find an encoder which maximizes the intra-class similarities and minimizes the inter-class similarities. To this end, we define the following loss functions:

$$\tilde{\ell}_{pos} = -\frac{1}{C} \sum_{k=1}^{C} \frac{1}{N_k} \sum_{i=1}^{N_k} \log \left[ \frac{1}{N_k - 1} \sum_{p=1, p \neq i}^{N_k} \exp \left( \mathcal{M}_{i,p}^{(k)} / \tau \right) \right] \tag{2}$$

$$\tilde{\ell}_{neg} = \frac{1}{C} \sum_{k=1}^{C} \frac{1}{N_k} \sum_{i=1}^{N_k} \log \left[ \frac{1}{\overline{N_k}} \sum_{n=1}^{\overline{N_k}} \exp \left( \mathcal{N}_{i,n}^{(k)} / \tau \right) \right] \tag{3}$$

where $C$ is the number of classes, and $\tau \in \mathcal{R}_+$ is a temperature parameter, and $\mathcal{M}_{i,p}^{(k)}$ and $\mathcal{N}_{i,n}^{(k)}$ denotes the $(i, p)$th element of $\mathcal{M}^{(k)}$ and the $(i, n)$th element of $\mathcal{N}^{(k)}$ respectively. The objective functions are defined as the expectation of the above (mini-batch) loss functions over the distribution of the data. The overall (single mini-batch) loss that we propose in this work is a weighted linear combination of the above conflicting objectives and the CE loss. When including the latter, the two objectives to minimize become:

$$\ell_1 = \lambda \tilde{\ell}_{pos} + (1 - \lambda)\mathcal{L}_{CE} \tag{4}$$

$$\ell_2 = \lambda \tilde{\ell}_{neg} + (1 - \lambda)\mathcal{L}_{CE} \tag{5}$$

where $\mathcal{L}_{CE}$ is the conventional cross-entropy loss, and $\lambda$ is a hyper-parameter that controls the weight of the SCL term in the objective function. The MOO is solved using both LS and exact Pareto optimal methods to fine-tuning the pretrained language model.

• **Linear Scalarization method**: this is the most straightforward approach to solve a MOO problem. It converts the latter to a single-objective optimization problem (Eq 6). Indeed, LS optimizes the weighted sum of the objectives, i.e.

$$\theta^* = \arg\min_{\boldsymbol{\theta}} \mathbb{E}_{(\boldsymbol{x}, y) \sim \mathcal{P}_D} \sum_{j=1}^{m} r_j \ell_j \tag{6}$$

where $\boldsymbol{r} \in \Omega^m$ is the preference vector, with

$$\Omega^m := \left\{ \boldsymbol{r} \in \mathbb{R}_+^m \,\middle|\, \sum_{j=1}^{m} r_j = 1, \text{and } r_j \geq 0 \;\forall j \right\}.$$

In our problem formulation $m = 2$. Although LS has some theoretical limitations, it has been shown to work well in practice.

• **Exact Pareto Optimal method**: it finds the intersection of the Pareto front with a given preference ray. This is achieved by considering the preference vector $\boldsymbol{r}$ as a ray in the loss space and training a neural network to reach a Pareto optimal solution on the inverse ray $\boldsymbol{r}^{-1}$. Thus, an Exact Pareto optimal (EPO) solution with respect to a preference vector $\boldsymbol{r}$ belongs to the set:

$$\mathcal{P}_{\boldsymbol{r}} = \left\{ \boldsymbol{\theta}^* \in \mathcal{P}_{\boldsymbol{\theta}} \,\middle|\, r_1 \ell_1^* = \cdots = r_m \ell_m^* \right\} \tag{7}$$

where $\ell_j^* = \ell_j(\boldsymbol{\theta}^*)$. This is achieved by balancing two goals: finding a descent direction towards the Pareto front and approaching the desired ray Mahapatra and Rajan [2021].

The pseudo-code of the proposed algorithm (using either LS or EPO) is described below.

---

**Algorithm 1** Fine-tuning strategy
___________________________________________________________________________________________________
$\boldsymbol{r} = [r_1, r_2]^T \in \Omega^2$ [6]The preference vector
$\tilde{\boldsymbol{\ell}} = [\tilde{\ell}_{pos}, \tilde{\ell}_{neg}]$ [7.5]The objective vector

  1: **while** not converged **do**
  2:     Sample a mini-batch for each class
  3:     **if** CE + LS-based **then**
  4:         $g_\theta \leftarrow \lambda(r_1 \nabla_\theta \tilde{\ell}_{pos} + r_2 \nabla_\theta \tilde{\ell}_{neg}) + (1 - \lambda)\nabla_\theta \mathcal{L}_{CE}$
  5:     **end if**
  6:     **if** CE + EPO-based **then**
  7:         $\boldsymbol{\beta} = [\beta_1, \beta_2] = EPO(f_\theta, \tilde{\boldsymbol{\ell}}, \boldsymbol{r})$
  8:         $g_\theta \leftarrow \lambda(\beta_1 \nabla_\theta \tilde{\ell}_{pos} + \beta_2 \nabla_\theta \tilde{\ell}_{neg}) + (1 - \lambda)\nabla_\theta \mathcal{L}_{CE}$
  9:     **end if**
10: **end while**
11: **return** $\theta$
___________________________________________________________________________________________________

## 4 Experiments

### 4.1 Datasets & Training Details

We evaluate our approach by measuring top-1 accuracy on the GLUE natural language understanding benchmark Wang et al. [2018]. The evaluation is measured on single sentence as well as sentence-pair text classification tasks. Table 1 summarizes each dataset based on their main task, domain, number of training examples, and number of classes.

| Corpus | #Train | #Test | #Classes | Task | Domain |
|--------|--------|-------|----------|------|--------|
| Single-Sentence Tasks | | | | | |
| CoLA | 8.5k | 1k | 2 | acceptability | linguistic publications |
| SST-2 | 67k | 1.8k | 2 | sentiment | movie reviews |
| Similarity and Paraphrase Tasks | | | | | |
| MRPC | 3.7k | 1.7k | 2 | paraphrase | news |
| Inference Tasks | | | | | |
| MNLI | 393k | 20k | 3 | NLI | multi-domain |
| QNLI | 105k | 5.4k | 2 | QA/NLI | Wikipedia |
| RTE | 2.5k | 3k | 2 | NLI | news/Wikipedia |

Table 1: Task descriptions and statistics

In our experiments, for few-shot learning setting, similar to Gunel et al. [2020], we sample 500 examples from the original validation set of each dataset to build our validation set, and half of the validation to build the test set. For full datasets training, we use the original validation set as our test set and we sample 10% from the original training set of the GLUE benchmark to build our validation set. In both settings, we run each experiment with 10 different seeds, and report the average test accuracy and the standard deviation along with the baselines. Best hyperparameters combination are picked based on the average validation accuracy.

We use the HuggingFace implementation of the pre-trained RoBERTa-base language model for all of our experiments. In this paper, we use RoBERTa-base due to the GPU RAM constraint. During all the fine-tuning runs, we use AdamW optimizer with a learning rate of 1e-5, batch size of 16, and dropout rate of 0.1. We optimized the temperature hyperparameter on the validation set by sweeping for $\tau \in \{0.1, 0.3, 0.5, 0.7, 0.9\}$, $\lambda \in \{0.1, 0.3, 0.5, 0.7, 0.9\}$, and $r_1 \in \{0.1, 0.3, 0.5\}$. Simulations show that models with best test accuracies across all experimental settings overwhelmingly use the hyperparameter combination $\tau = 0.3$, $\lambda = 0.3$, and $\boldsymbol{r} = [0.1, 0.9]$

## 5   Results & Analysis

Here, we report the obtained results of our approach on the few-shot learning setting (20, 50, 100 annotated examples) with those obtained by the baselines that fine-tune the RoBERTa-base with CE and CE + SCL loss, respectively. Performance is measured in terms of the Accuracy metric on the test set. We run each experiment with 10 different seeds (details of the experimental setup are explained in Experiments section). For both CE + LS-based and CE + EPO-based losses, experiments were carried out using different preference vectors (See Ablation study for more details). As shown in Table 2, we observe that our approach obtains significantly better performance than the baselines. For instance, the CE + EPO-based loss achieves, on SST-2, an accuracy of 68.96%, which is a 9.24 and 8.6 points improvement over CE and CE + SCL respectively when the number of annotated examples is 20. Similarly, CE + LS-based objective achieves better results than the baselines when the RoBERTa-base is fine-tuned using 20 labeled examples, with 1.85 points improvement on QNLI and 1.9 points improvement on MNLI. This shows that our approach *generalizes* better than the baselines. We believe that this is due to the fact that good generalization (high-quality representations) requires capturing well the similarity between examples in one class and contrasting them with examples from other classes. Note that, as we increase the number of annotated examples, performance improvement over the baseline decreases, leading to 1.5 points improvement on SST-2 for 50 examples and 0.23 points improvement for 100 examples. However, our CE + EPO-based method achieves consistent improvements on MNLI dataset across all data regimes. On QNLI dataset, we see that our method is outperformed by the CE + SCL when the number of labeled examples is 50. However, overall, our method performs better than the baselines in the few-shot learning setting.

Figure 4 shows the tSNE plots of the learned sentence embeddings on SST-2 test set when RoBERTa-base is fine-tuned using only 20 annotated examples with CE + LS-based, CE + EPO-based, and CE + SCL losses. As we can clearly see, our approach forces the encoder to better separate the classes in the embedding space, while forcing it to achieve more compact clustering.

| Loss | N | SST-2 | QNLI | MNLI |
|------|---|-------|------|------|
| CE | 20 | 59.72 ± 5.9 | 50.57 ± 1.8 | 34.07 ± 1.5 |
| CE + SCL | 20 | 60.34 ± 5.3 | 50.80 ± 1.8 | 33.25 ± 1.5 |
| CE + LS-based (Ours) | 20 | **72.06 ± 5.4** | **51.89 ± 1.6** | **34.81 ± 1.1** |
| CE + EPO-based (Ours) | 20 | **68.96 ± 3.5** | 51.44 ± 1.2 | 33.94 ± 0.9 |
| CE | 50 | 81.64 ± 1.6 | 61.20 ± 2.5 | 37.36 ± 2.1 |
| CE + SCL | 50 | 81.42 ± 4.8 | **66.02 ± 2.4** | 38.20 ± 1.1 |
| CE + LS-based (Ours) | 50 | **81.92 ± 1.6** | 62.06 ± 4.2 | **39.55 ± 1.8** |
| CE + EPO-based (Ours) | 50 | **82.93 ± 0.9** | 62.93 ± 1.6 | **40.52 ± 1.5** |
| CE | 100 | 85.32 ± 1.4 | 72.74 ± 1.4 | 44.87 ± 1.5 |
| CE + SCL | 100 | 85.41 ± 1.3 | 73.37 ± 1.1 | 43.79 ± 1.9 |
| CE + LS-based (Ours) | 100 | **85.64 ± 1.4** | 73.39 ± 1.2 | **46.51 ± 1.7** |
| CE + EPO-based (Ours) | 100 | **85.43 ± 1.2** | **74.29 ± 0.8** | **46.89 ± 1.6** |

Table 2: Few-shot classification accuracies on test sets of the GLUE benchmark where we have N=20, 50, 100 annotated examples for fine-tuning RoBERTa-base model. We report mean (and standard deviation) performance over 10 different seeds for each experiment.

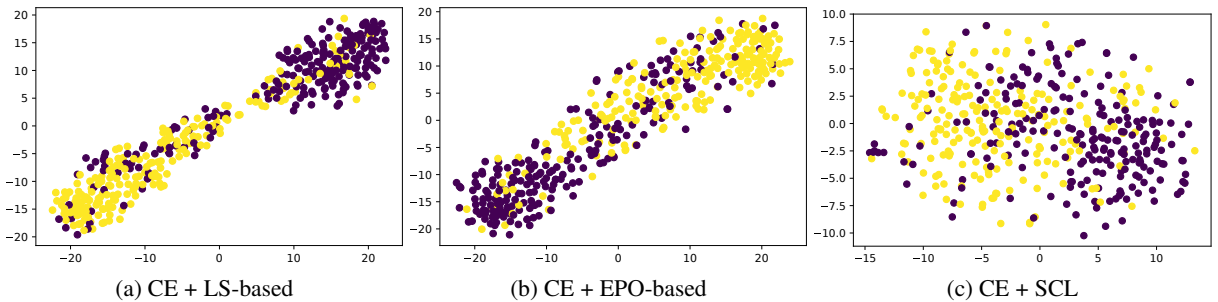

(a) CE + LS-based          (b) CE + EPO-based          (c) CE + SCL

Figure 4: T-SNE plot of the learned embeddings on the SST-2 test set where we have 20 annotated examples to fine-tune RoBERTa-base language model fine-tuned with CE + LS-based (left), with CE + EPO-based (mid) and with CE + SCL objective (right).

In Table 3, we report accuracy measurements on the test set (the entire original validation set) using our approach on downstream tasks from the GLUE benchmark using full dataset for fine-tuning RoBERTa-base. As we can see, our approach yields better results on several datasets. For instance, our CE + EPO-based loss achieves, on CoLA, an accuracy of 83.65 which is 0.63 point improvement over the CE + SCL loss. However, we also see that the baselines can achieve better results than our approach on some datasets.

## 6 Ablation Study

We performed extensive ablations for 3 important hyperparameters namely, the preference vector, the temperature scalar and the weighting hyperparameter $\lambda$ on SST-2 dataset, when the number of labeled examples is 50.

In this paper, we treat the preference vector as a hyperparameter to be fine-tuned. Thus, an appropriate preference vector can help the model to learn more powerful and generalizable representations for the downstream tasks. To this end, we run several simulations, each has a specific preference vector reflecting a different trade-off between the two objectives. Experiments show that the minimization of the inter-class similarity-based objective ($\tilde{\ell}_{neg}$ through $\ell_2$) is crucial in order to get high-quality embeddings as can be seen in figure 5a. We also investigated the effect of varying the hyperparameter $\tau$ on the acuuracy on the validation set. As we can see in figure 5b, changing the temperature scalar leads to small changes on the accuracy score which means that our approach is more stable with respect to the temperature hyperparameter. We fine-tuned the model with different values of $\lambda$ and evaluated it on the validation set. As shown in Figure 5c the model performs well when the value of $\lambda$ is small, unlike the CE + SCL loss, which means that our method relies on the cross-entropy loss for learning good representations.

| Loss | SST-2 | CoLA | MRPC | RTE | QNLI | MNLI | avg |
|------|-------|------|------|-----|------|------|-----|
| CE | 93.12 ± 0.4 | 82.73 ± 0.6 | **88.23**± **0.9** | 75.74 ± 1.0 | 86.87 ± 1.1 | 84.29 ± 0.6 | 85.16 |
| CE + SCL | 93.51 ± 0.5 | 83.02± 0.6 | 87.59 ± 0.7 | **77.68**± **1.4** | 88.53 ± 0.9 | 84.91 ± 0.5 | 85.87 |
| CE + LS-based (Ours) | **94.01 ± 0.8** | **83.49 ± 0.7** | 87.10 ± 0.9 | 75.88 ± 1.2 | 88.15 ± 1.4 | **85.09 ± 0.5** | 85.62 |
| CE + EPO-based (Ours) | **94.18 ± 1.3** | **83.65 ± 0.9** | 87.05 ± 0.4 | 76.93 ± 1.3 | 88.49 ± 1.8 | **85.19 ± 0.5** | 85.92 |

Table 3: Accuracy score no the validation set of the GLUE benchmark. We compare fine-tuning RoBERTa-base with CE, CE + SCL, CE + LS-based and CE + EPO-based objectives. We report average accuracy (and standard deviation) across 10 different seeds.

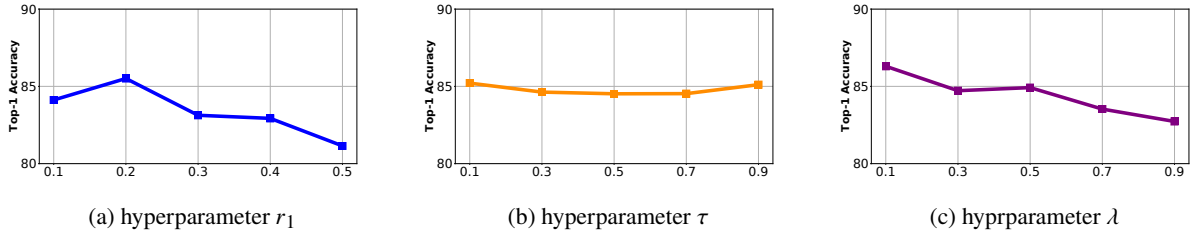

(a) hyperparameter $r_1$                     (b) hyperparameter $\tau$                     (c) hyprparameter $\lambda$

Figure 5: Evaluation on the impact of scaling parameters $r_1$ (left) where $r_1 + r_1 = 1$, temperature $\tau$ (mid) and $\lambda$ (right), all measured on SST-2 validation set, where we have 50 labeled examples.

## 7 Conclusion

In this paper, we propose a novel learning strategy for text classification tasks. We formulate the supervised contrastive learning problem as a Multi-Objective Optimization problem. The proposed loss function includes both supervised contrastive learning loss and the conventional cross-entropy loss. To solve the optimization problem, we employed two well-known approaches, namely the linear scalarization and the exact Pareto optimal solution search method. We evaluated the proposed method in few-shot learning (20, 50, 100 labeled examples) as well as the full dataset training on several datasets from GLUE benchmark. Empirically, we demonstrate the superior performance of our solution over two competing approaches for fine-tuning RoBERTa-base model. As a future work, we aim to adapt the proposed method for the self-supervised learning setting. We will also extend our approach to different application domains such as computer vision and graph representation learning.